\pdfoutput=1

\documentclass[11pt,table,xcdraw]{article}

\usepackage{acl}

\usepackage{xcolor} 
\usepackage{times}
\usepackage{latexsym}
\usepackage{graphicx}
\usepackage{booktabs}
\usepackage{amsmath}
\usepackage{amsfonts}
\usepackage{setspace} 
\usepackage{url}
\usepackage[vlined]{algorithm2e}
\usepackage{xspace}

\usepackage[T1]{fontenc}

\usepackage[utf8]{inputenc}

\usepackage{microtype}
\usepackage{silence}  
\title{Concept-aware Training \\Improves In-context Learning Ability of Language Models}

\author{Michal Štefánik \and Marek Kadlčík \\
  Faculty of Informatics, Masaryk University, Czech Republic \\
  \texttt{\{stefanik.m,kadlcik\}@mail.muni.cz}
\vspace*{-2\baselineskip}}

\begin{document}
\maketitle
\begin{abstract}

Many recent language models (LMs) of Transformers family exhibit so-called in-context learning (ICL) ability, manifested in the LMs' ability to modulate their function by a task described in a natural language input. Previous work curating these models assumes that ICL emerges from vast over-parametrization or the scale of multi-task training. However, a complementary branch of recent theoretical work attributes ICL emergence to specific properties of training data and creates functional in-context learners in small-scale, synthetic settings.

Inspired by recent findings on data properties driving the emergence of ICL, we propose a method to create LMs able to better utilize the in-context information, by constructing training scenarios where it is beneficial for the LM to capture the \textbf{analogical reasoning concepts}. We measure that data sampling of Concept-aware Training (\textsc{CoAT}) consistently improves models' reasoning ability. As a result, the in-context learners trained with \textsc{CoAT} on only two datasets of a single (QA) task perform comparably to larger models trained on 1600+ tasks.

\end{abstract}

\section{Introduction}
\label{sec:intro}

\begin{figure}[tbh]
    \centering
    \includegraphics[width=0.5\textwidth]{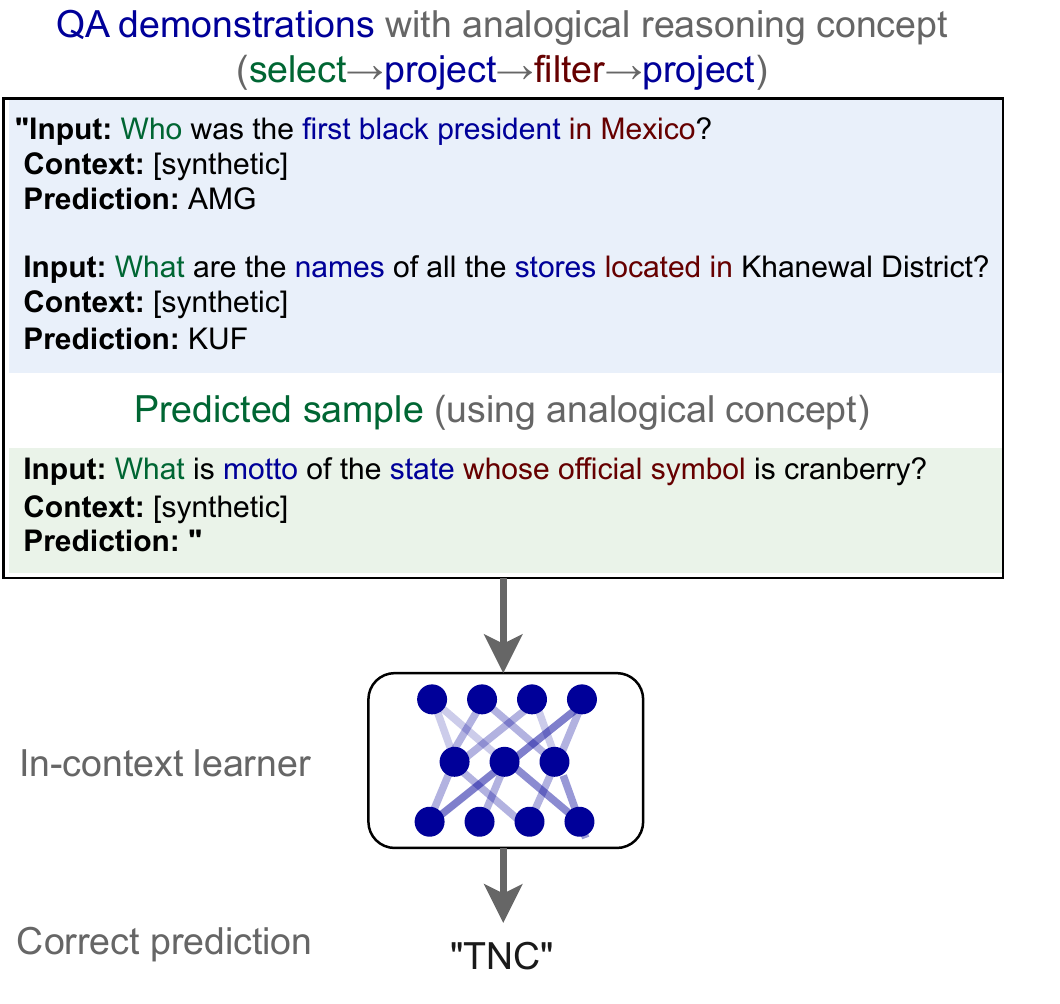}
    \caption{We propose a Concept-aware Training (\textsc{CoAT}) method to train the model for ability to capture reasoning concepts presented in given demonstrations (§\ref{sec:training}). We show that pre-training using \textsc{CoAT} largely improves models' performance compared to random demonstrations' selection and results in a quality similar to large-scale multi-task learning when training on only two QA datasets.}
    \label{fig:abstract}
\end{figure}

The in-context learning (ICL), as initially uncovered by \citet{gpt3}, is a task requiring language models (LMs) to infer and apply correct functional relationship between inputs and outputs (i.e. \textit{demonstrations}) presented in user-given input prompt \citep{li2023transformers}. 
Given that a set of demonstrations can be obtained for any machine learning task, in-context learning models present a much more versatile and very practical alternative to task-specific models. 

Modern in-context learners can often perform ICL with quality comparable to task-specialized models \citep{Zhao2023_LLMSurvey,stefanik-etal-2023-resources}. 
However, it remains unclear why some LMs are able of ICL in applicable quality, while others are not;
Initial work introducing GPT3 \citep{gpt3} followed by \citet{thoppilan2022lamda,chowdhery2022palm}; \textit{inter alia} explain ICL as an emergent consequence of models' scale. 
But more recent LMs \citep{sanh2022multitask,wang-etal-2022-super,flan,ouyang2022training} are based on 10 to 100 times smaller models while reaching comparable ICL quality. These works claim the ICL ability to a vast volume and diversity of pre-training tasks and instruction formats, allowing the model to generalize.
Is this sufficient evidence that \textbf{the in-context learning ability is \textit{caused} by the scale of data or model size?}

Complementary branch of theoretical studies is more specific in identifying covariates responsible for the emergence of ICL in \textbf{data irregularities}, i.e. the properties of the data that can not be explained by mere statistical co-occurrence of tokens. Notably, \citet{chan2022data} find such property in a statistical \textit{burstiness}, characterized by specific concepts co-occurring in clusters, conditionally to the context. \citet{xie2022an} identify the key property in the occurrence of long-range text dependencies that can be resolved by identifying \textit{latent} \textit{concepts} that underpin these dependencies. Both these works demonstrate the emergence of ICL in-silica by creating a small synthetic dataset exhibiting given property and fostering ICL ability when used in training, even with small models and data.

In this work, we adapt and empirically verify the recent theories revolving around the data irregularities shown to foster ICL in synthetic settings; We propose a pre-training task (§\ref{sec:training}) where we \textit{encourage} concept-dependent irregularity in data, and hence, the task requires the model to in-context \textit{learn to exploit} the reasoning concepts that explain these irregularities (Fig.~\ref{fig:abstract}). We refer to this approach as \textbf{Concept-aware Training} (\textbf{CoAT}).

We explore the impact of this adjustment in controlled settings (§\ref{sec:results}) and find that CoAT consistently improves models' performance on reasoning tasks of SuperGLUE \cite{wang2019superglue}, in some cases improving the accuracy of commonly-used random data selection 2--3-times. Consequentially, CoAT pre-training allows in-context learners trained in relatively small data settings of merely two (2) QA tasks to reach the accuracy of in-context few-shot learning that is similar or even superior to equal-size, or larger models trained on 1,600+ tasks.


\section{Background}
\label{sec:background}

\paragraph{Methods for training in-context learners}

In-context learning ability, including few-shot ICL, was first uncovered in \textsc{GPT3} \cite{gpt3} trained unsupervisedly for causal language modelling. With no substantial methodical differences to its predecessors \citep{Radford2018gpt,gpt2} the emergence of ICL was attributed to GPT3's scale, having grown to over 170-billion parameters since \textsc{GPT2} (appx. 800M params). 

Not long after, a pivotal work of \citet{pet} on a pattern-exploiting training (PET) has shown that even much smaller (110M) encoder models like BERT \cite{devlin-etal-2019-bert} can be fine-tuned with few-shot demonstrations using self-training to similar accuracy, first disputing the initial assumption on the necessity of the scale in learning unseen tasks.

A new branch of autoregressive generation models soon further undermined the assumption of the size conditioning of ICL. The work of \citet{min-etal-2022-metaicl} is, to our knowledge, the first to fine-tune smaller pre-trained models (<1B parameters) on a large mixture of tasks in the few-shot prompt format, showing that such models are also able to perform well on previously unseen tasks, in the ICL setup. Following approaches also train smaller models in instruction tuning \cite{sanh2022multitask,wang-etal-2022-super} on large mixtures of tasks, assuming that the model's ability to accurately learn an unseen task without updates emerges from a large variety of diverse instruction formats and task types. A recently popularised approach of \textsc{InstructGPT} \cite{ouyang2022training} also presents an adaptation of the instruction-tuning approach, fine-tuning the model over a large variety of automatically annotated user instructions.

Recently, the instruction-tuning approach was complemented by Chain-of-Though (CoT) generation objective \cite{wei2022chain}, where the model is trained to augment the response with a sequence of natural-language steps deducing its answer \citep{Zhao2023_LLMSurvey}, adopted, for instance, by \textsc{Flan} models \cite{chung2022_flan}, or by joint training on a programming code \cite{chen2021evaluating}. Both these extensions were empirically shown to enhance ICL of instruction-tuned models \cite{fu2022gptroadmap}.









\paragraph{Analyses of ICL} Despite surprising accuracy of in-context learning of recent LMs, it remains a matter of open discussion as \textit{why} the in-context learning emerges. 

Recent studies shed some light in this direction through controlled experimentation, finding that the LMs' decision-making in ICL does not align with human intuition; 
Notably, \citet{lu-etal-2022-fantastically} first report on the sensitivity of LMs to the specific formulation of the instructions in the prompt, while \citet{liu-etal-2022-makes} report on LMs' surprising sensitivity to the ordering of in-context demonstrations. 
Further, it was shown that LMs perform ICL comparably well when the labels of the demonstrations are randomly shuffled \citep{what-makes-incontext-work} or when the presented CoT sequences do not make sense \citep{wang2023towards}.
These phenomena might be explained by LMs' reliance on the semantics of the labels from the pre-training \cite{wei2023larger}; Note how such behaviour differs from learning a functional relation of inputs and labels, that we might expect from in-context learners \cite{li2023transformers}.

Other studies report that under the right conditions, LMs are able to foster the ability learn to identify functional relationships solely from the input; For instance, studies of \citet{akyurek2023what,Li2023TransformersAA} show that it is possible to pre-train Transformers to learn regression functions solely from input context accurately. 

\citet{xie2022an} might be the first to identify the causal effects of ICL in specific data properties, rather than data scale, identifying the causal of the ICL in the presence of the latent concepts that LMs need to utilise to improve in the training task (either pre-training or fine-tuning). Related work attributes ICL to similar data irregularities, such as \textit{burstiness} \cite{chan2022data} or \textit{compositionality} \citep{hahn2023theory}. Note that these studies are not conflicting with the aforementioned empirical results, but rather explain the causes of their success; For instance, in multi-task training, LMs might indeed inherently learn to identify shared concepts from inputs \citep{wies2023learnability}.

Our work builds upon these findings, but contrary to previous work, we implement the idea of concept-aware training in real-world settings, i.e. with publicly-available datasets and commonly-used pre-trained models. Further, we measure the impact of our approach in extrinsic evaluation over ten reasoning tasks of SuperGLUE \cite{wang2019superglue} and compare it to the previous work utilising magnitudes of more compute in multi-task training.






\section{Concept-aware Training (\textsc{CoAT})}
\label{sec:training}

Concept-aware Training (\textsc{CoAT}) method adapts the findings of previous work in data-driven emergence of ICL in a conditional selection of few-shot demonstrations presented in the training prompts. For this purpose, we adopt the format of training prompts widely used in the previous work (as instantiated in Appx.~\ref{appx:evaluation}), including $k$ demonstrations consisting of the inputs $x$ and associated labels $y$:
\begin{equation*}
    [x_1, y_1, <\!\!sep\!\!>, ..., x_k, y_k, <\!\!sep\!\!>, x_\text{pred}] \rightarrow y_\text{pred}    
\end{equation*}
The main condition of \textsc{CoAT} is to pick demonstrations that present a reasoning concept $\mathcal{C}$ that is shared with the predicted example $(x_\text{pred}, y_\text{pred})$, thus making the demonstrations \textbf{informative} for the correct prediction. In such settings, it is beneficial for the trained model to learn to \textit{extract} and \textit{apply} concepts presented in the input prompt.

However, as the sole \textit{informativeness} condition may pick demonstrations very similar to the predicted input, we further propose a \textbf{non-triviality} condition to filter the demonstrations to ones for which it is `difficult' for the model to respond correctly. This also enhances the heterogeneity of the concepts that co-occur among the demonstrations and may avoid over-reliance on the presence of a small set of specific concepts in small-data regimes.

\section{Experiments}
\label{sec:experiments}

\subsection{Implementation of \textsc{CoAT}}

We instantiate the \textsc{CoAT} method in two training stages: First, by we train LM on a synthetic QA dataset with explicitly annotated reasoning concepts. Second, we refresh the LM's ability to work with natural-language prompts by further tuning on a chosen organic QA dataset. In a result, our models are trained on only two QA datasets.

\paragraph{Informativeness condition} 

For the first step, we find a large collection of annotated reasoning concepts in a TeaBReaC dataset of \citet{trivedi-etal-2022-teaching}, containing more than 900 unique explanations over a relatively large set of \textit{synthetic} QA contexts. Each explanation maps a natural question to the answer span through a sequence of declarative reasoning steps, such as ``select$\rightarrow$group$\rightarrow$project''. We use these patterns as informative concepts $\mathcal{C}$ and construct training input texts from demonstrations sharing the concept with the predicted sample (Fig.~\ref{fig:abstract}).

Second, to restore the model's ability to utilise the semantics of a natural language, in the following step, we fit the resulting model to \textit{natural} inputs by further fine-tuning on AdversarialQA dataset \citep{bartolo-etal-2021-improving}; As we do not find the reasoning concepts annotated among the available QA datasets, in this case, we naively use the initial word of the question (``Who'', ``Where'', ...) as the shared concept, aware that such-grouped demonstrations are often not mutually informative.

\paragraph{Non-triviality condition} 

We implement the \textit{non-triviality condition} of \textsc{CoAT} by (i) selecting a random set of samples $X_\text{info}: |X_\text{info}| = 20$ from the demonstrations that pass the \textit{Informativeness} condition. (ii) then we iteratively pick a sequence of $i \in 1...k$ demonstrations from this set, with a randomly-chosen $k: 2\leq k \leq 8$.
\begin{enumerate}
    \item For each sample $(x_j, y_j) \in X_\text{info}$ in the set, we compute a likelihood of generating the correct prediction, if a given sample is included among demonstrations. This likelihood is computed as a product of likelihoods of generating correct prediction $y_\text{pred}$ in the teacher-forced generation.
    \item In each step $i$, we include among the demonstrations a sample for which the likelihood of generating correct prediction is \textit{minimal}.
\end{enumerate}




\subsection{Training Setup of CoAT}
\label{sec:training_setup}

To match the architectures of the previous work, we fine-tune \textsc{Tk-CoAT} from \textsc{mT5-Large} of \citet{xue-etal-2021-mt5} on (1) TeaBReaC dataset, followed by (2) AdversarialQA dataset. In both steps, we fine-tune all model parameters for teacher-forced next-token prediction, conventionally used in training sequence-to-sequence language models. For both training stages, we use a learning rate = $2e^{-5}$, effective batch size = 30 samples, and early stopping based on evaluation loss on a standardized validation set of each dataset. Other parameters of training configuration default to Training Arguments of Transformers library \cite{Wolf2019HuggingFacesTS} in version 4.25.1.
For readability, we implement\footnote{Our implementation of \textsc{CoAT} with resulting models can be found on: \url{https://github.com/MIR-MU/CoAT}} the relatively complex demonstrations' selection as a new objective of Adaptor library~\cite{stefanik-etal-2022-adaptor}.
The picked demonstrations are encoded in the format consistent with the evaluation (Appendix~\ref{appx:evaluation}). 


\subsection{Baselines}

\paragraph{Random demonstrations choice (\textsc{Tk-random})}
We assess the impact of the controlled selection of demonstrations against a baseline trained in the same settings but picking the in-context demonstrations \textit{randomly} with uniform probability over the whole training set. Analogically to \textsc{CoAT}, the baseline is also trained sequentially on two datasets (TeaBReAC + AdversarialQA), and all other training parameters remain unchanged.

\paragraph{Demonstrations choice without non-triviality (\textsc{Tk-info})}
Additionally, we perform ablation of the Non-triviality condition introduced in §\ref{sec:training} by picking the demonstrations passing only the \textit{Informativeness} condition. Hence, such-picked demonstrations in the training input context are mutually informative by the shared concept but can exhibit cases where some of the demonstrations are very similar to the predicted sample, making it trivial for the model to perform correct prediction. Again, all other training settings remain unchanged (§\ref{sec:training_setup}).

\begin{table*}[ht]
\vspace*{2mm}
\centering
\scalebox{0.85}{
\begin{tabular}{lcccccc}
\hline
 & \multicolumn{1}{l}{\textsc{Tk-random-1B}} & \multicolumn{1}{l}{\textsc{Tk-info-1B}} & \multicolumn{1}{l}{\textsc{Tk-CoAT-1B}} & \multicolumn{1}{l}{\textsc{T0-3B}} & \multicolumn{1}{l}{\textsc{Tk-Instruct-1B}} & \multicolumn{1}{l}{\textsc{T5-Flan-1B}} \\ \hline
AxG & 48.6 & 50.0 & 50.2 & 64.8 & 51.9 & \underline{83.4} \\ 
\rowcolor[HTML]{ECF4FF} 
AX-b & 43.0 & 42.6 & 54.0 & 35.2 & \underline{57.2} & 25.4 \\
WSC & 53.2 & 52.0 & 53.2 & 53.6 & 50.2 & \underline{71.1} \\
\rowcolor[HTML]{ECF4FF} 
CB & 22.1 & 47.2 & 46.4 & \underline{48.0} & 46.0 & \ \,92.0$^*$ \\
RTE & 15.2 & 49.2 & 53.4 & 51.0 & \underline{55.6} & \ \,91.8$^*$ \\
\rowcolor[HTML]{ECF4FF} 
WiC & 18.6 & 53.2 & 53.2 & 53.2 & \underline{54.0} & \ \,68.6$^*$ \\
ReCoRD & 14.3 & 15.5 & 15.7 & \underline{20.5} & 12.8 & \ \,40.6$^*$ \\ 
\rowcolor[HTML]{ECF4FF} 
BoolQ & 34.1 & 19.6 & \underline{63.4} & 56.0 & \ \,63.4$^*$ & \ \,90.6$^*$ \\
COPA & 73.9 & 61.5 & \underline{75.7} & 56.8 & \ \,76.9$^*$ & \ \,97.1$^*$ \\
\rowcolor[HTML]{ECF4FF} 
MultiRC & \ 5.2 & 3.8 & 11.4 & \underline{56.6} & \ \,62.6$^*$ & \ \,88.4$^*$ \\ \hline
\end{tabular}

}
\caption{\textbf{SuperGLUE accuracy} of ICL models evaluated in few-shot (k=3) settings. Reported results utilize the best-performing Promptsource template for each task+model. In cases marked with $^*$, the task was used in the model's training; \underline{Underlined} are the best results per each unseen task.
}
\label{table:superglue}
\end{table*}

\section{Results and Discussion}
\label{sec:results}

Table~\ref{table:superglue} evaluates the reasoning ability of in-context learner trained using CoAT (\textsc{Tk-CoAT}; §\ref{sec:training}), as compared to using contexts with randomly-drawn demonstrations used by previous work (\textsc{Tk-random}). We observe that concept-aware construction of training samples improves prediction quality in all cases except for the Winograd Schema Challenge (WSC), where it performs on par. However, in some tasks (e.g. RTE or WiC), CoAT enables the in-context learning of tasks where the initial model does not exceed random performance, with measured accuracy gains exceeding 300\%.

Ablation of the non-trivial filtering presented in the comparison of \textsc{Tk-CoAT} with \textsc{Tk-info} shows that selective sampling of demonstrations is important for in-context learning of 5 out of 10 SuperGLUE tasks, whilst, for the others, the results are comparable.

Finally, a comparison of \textsc{Tk-CoAT} trained on two tasks of the same (QA) type with in-context learners trained on 35--1,600+ tasks of the same, or larger size shows only small differences in accuracy. For instance, the 1-billion-parameter \textsc{Tk-CoAT} performs better than the 3-billion \textsc{T0} in 5 cases and comparably in another 3 cases (WSC, CB and WiC). A comparison with \textsc{Tk-instruct} of the same size shows similar scores in all the cases, even in two out of three tasks that \textsc{Tk-instruct} was trained on. 





\section{Conclusion}
\label{sec:conclusion}

This work introduces a Concept-aware Training (\textsc{CoAT}) method; Building upon the recent findings in theories of the emergence of in-context learning, \textsc{CoAT} proposes to pre-train models on data that manifest data irregularities shown to give rise to in-context learning ability.

We implement this approach in constructing training prompts presenting demonstrations that share analogical reasoning concept with the predicted sample, allowing the trained in-context learner to benefit from learning to extract the reasoning concept that explains the target prediction.

We find that \textsc{CoAT} can help to foster more efficient in-context learning ability as demonstrated by a comparison with a random selection of in-context demonstrations commonly adopted by previous work in instruction tuning. 
Finally, \textsc{CoAT} shows its efficiency in delivering performance comparable to models trained on over 1,600 tasks, while utilising only two datasets of the same task type. 
This shows that further scaling of the model size and task collections, both widely adopted in previous work, might not be necessary for enhancing new-task learning ability. 
Instead, other design choices, such as demonstration selection, may provide similar gains for a substantially smaller computing price.


\section*{Ethical Considerations \& Broader Impact}
\label{sec:ethical}

The initial motivation of our work is to minimise the environmental impact of developing new models through the minimisation of hardware requirements. We hope to motivate the developers of future in-context learners to explore other aspects related to resulting ICL quality, outside data and model size scaling.

We also note that data-efficient methods for training ICLs might open possibilities for creating accurate ICLs for languages outside English, where training data is scarce. We look forward to exploring the possibility of using a single QA dataset in the target language for creating target-language in-context learner(s).

\bibliography{marekov,stefanik}
\bibliographystyle{acl_natbib}

\appendix

\section{Training and evaluation prompt format}
\label{appx:evaluation}

We train and evaluated all models over both datasets and demonstrations selection strategies consistently using a randomly-chosen number of demonstrations: $2 \leq k \leq 8$ and contexts constructed in the following format:
\bigskip

\noindent \textit{``Input: $x_1$ Prediction: $Y_1$ \\
Input: $x_2$ Prediction: $Y_2$ \\
Input: $x_3$ Prediction: $Y_3$ \\
Input: $x_\text{pred}$''}
\bigskip

\noindent with the generation label $Y_3$.

\section{Computational Requirements}

We run both training and evaluation experiments using four \textsc{NVidia A100-SXM-80GB}, but we release the reproduction scripts that should run on a single 80GB GPU or even much smaller-memory GPUs with smaller models. The training in total takes 117h for \textsc{Tk-CoAT}, and 92h in total for \textsc{Tk-QA-random} to converge.

The time and computational requirements of evaluation depend largely on the size of the evaluated model; We are able to evaluate the models up to 3B parameters on a single \textsc{NVidia A100-SXM-80GB}. However, larger models require multiple GPUs. The evaluation of Concept Few-shot learning on all our datasets, together with the Random reference evaluation takes approximately 2~hours for a 1B model.

\end{document}